# SAVeD: A First-Person Social Media Video Dataset for ADAS-equipped vehicle Near-Miss and Crash Event Analyses

**Shaoyan Zhai, PhD student**
SST, Department of Civil, Environmental and Construction Engineering
University of Central Florida, Orlando, FL 32816, United States
Email: shaoyan.zhai@ucf.edu

**Mohamed Abdel-Aty, Professor**
SST, Department of Civil, Environmental and Construction Engineering
University of Central Florida, Orlando, FL 32816, United States
Email: m.aty@ucf.edu

**Chenzhu Wang, Postdoctoral Researcher\***
**Corresponding author**
SST, Department of Civil, Environmental and Construction Engineering
University of Central Florida, Orlando, FL 32816, United States
Email: chenzhu.wang@ucf.edu

**Rodrigo Vena Garcia, PhD student**
SST, Department of Civil, Environmental and Construction Engineering
University of Central Florida, Orlando, FL 32816, United States
Email: rodrigo.venagarcia@ucf.edu

**Abstract**

The advancement of safety-critical research in driving behavior in ADAS-equipped vehicles require real-world datasets that not only include diverse traffic scenarios but also capture high-risk edge cases such as near-miss events and system failures. However, existing datasets are largely limited to either simulated environments or human-driven vehicle data, lacking authentic ADAS (Advanced Driver Assistance System) vehicle behavior under risk conditions. To address this gap, this paper introduces SAVeD, a large-scale video dataset curated from publicly available social media content, explicitly focused on ADAS vehicle-related crashes, near-miss incidents, and disengagements. SAVeD features 2,119 first-person videos, capturing ADAS vehicle operations in diverse locations, lighting conditions, and weather scenarios. The dataset includes video frame-level annotations for collisions, evasive maneuvers, and disengagements, enabling analysis of both perception and decision-making failures. We demonstrate SAVeD's utility through multiple analyses and contributions: (1) We propose a novel framework integrating semantic segmentation and monocular depth estimation to compute real-time Time-to-Collision (TTC) for dynamic objects. (2) We utilize the Generalized Extreme Value (GEV) distribution to model and quantify the extreme risk in crash and near-miss events across different roadway types. (3) We establish benchmarks for state-of-the-art VLLMs (VideoLLaMA2 and InternVL2.5 HiCo R16), showing that SAVeD's detailed annotations significantly enhance model performance through domain adaptation in complex near-miss scenarios. Compared to existing resources, SAVeD offers significantly more high-risk scenes, facilitating the development of advanced prediction models and vision-language understanding. With ongoing expansion, SAVeD is positioned to become a foundational resource for safety, VLLM, and human-centered ADAS vehicle research.

## Background & Summary

### *Current Research on Autonomous Vehicle crashes*

With the rapid deployment of Autonomous Vehicles (AVs) on public roads, growing concerns and interests are emerging in systematically evaluating their safety performance in real-world environments. Although existing large-scale autonomous driving datasets such as Cityscapes (*1*), the Waymo Open Dataset (*2*), nuScenes (*3*), and Argoverse (*4*) have significantly advanced research in perception and path planning, these datasets contain very limited or no publicly available scenes involving real-world autonomous vehicles engaged in traffic crashes. On the other hand, video crash datasets like Nexar Dashcam Collision Prediction Dataset (*5*), DADA-2000 (*6*), CADP (*7*), MM-AU (*8*), HDD (*9*), and DOTA (*10*), which are widely used for driving behavior analysis and crash prediction, primarily consist of footage from human-driven vehicles or third-person views captured by urban surveillance cameras. Consequently, the real-world performance of autonomous vehicles during crash scenarios is not accurately captured.

Moreover, it is important to highlight that datasets such as the CA DMV (*11*) disengagement reports and AVOID (*12*) primarily consist of textual crash reports related to AVs, although these text-based datasets provide high credibility and are widely used in official reporting, they lack the temporal and spatial richness needed to analyze real-time AV behavior prior to incidents, thereby limiting their usefulness for detailed behavioral analysis or simulation-based reconstruction. lacking synchronized video evidence. These reports typically provide only brief description of incidents or system disengagements, which limits the ability to capture the full temporal dynamics and environmental context of crashes involving AVs.

To address this critical data gap, SAVeD (Social media-based ADAS-equipped Vehicle event Dataset) offers one of the first large-scale, publicly available video datasets explicitly addressing rear-world ADAS-equipped vehicle crashes and near-miss events. SAVeD integrates first-person view videos from multiple sources, all clearly involving autonomous vehicle participation. It features fine-grained frame-level annotations including the precise frame of crash occurrence, the moment ADAS-equipped vehicle initiated evasive maneuvers, and detailed contextual information such as surrounding traffic conditions, responsible parties, and concise crash summaries. This dataset lays a novel foundation for comprehensive studies on ADAS-equipped vehicle reaction time analysis, fault diagnosis, pre- crash behavior modeling, perception, and decision-making, providing essential support for advancing ADAS-equipped vehicle safety evaluation and improvement in real-world conditions.

**Table 1** provides a comparative overview of several publicly known ADAS-equipped vehicle crash datasets, focusing on their geographic coverage, inclusion of crash versus near-miss events, availability of descriptive reports, and dataset scale. While official datasets such as those from NHTSA and CA DMV remain primary sources, they are limited to crash-only incidents and typically provide only structured text-based reports. Proprietary datasets like Waymo add value by including video data, yet their access is often restricted.

Automated driving technology is progressing at an unprecedented pace, accompanied by high expectations for improving traffic efficiency and reducing crashes caused by human error. Among various automation levels, this study focuses on vehicles equipped with automated a widely deployed but semi-automated class of vehicle automation requiring continuous human supervision and occasional manual intervention. Despite the increasing prevalence of ADAS-equipped vehicles on public roads, traffic crashes and safety concerns related to these systems remain, raising important questions about their reliability and operational performance in real-world environments (*13, 14*). Consequently, systematic evaluation of ADAS vehicle performance within actual traffic scenarios has become a pivotal research focus within intelligent transportation systems.

**TABLE 1 Comparison of AV events Datasets in Terms of Scope, Content, and Accessibility**

| Name | location | Crash Only / With Near Miss | Text Report / With video | Sample size |
|------|----------|------------------------------|---------------------------|-------------|
| Avoid(*12*) | Global | Crash only | text report only | 1000+ |
| CA DMV's dataset(*11*) | California | Crash only | text report only | 821(Until May 2025) |
| WOD-E2E (*14*) | USA | Near Miss only | Video only | 4021 |
| NHTSA's dataset(*15*) | USA | Crash only | text report only | 6000+(Until April 2025) |
| Deep accident(*16*) | Unreal | Unreal | Unreal | 57K annotated frames |
| SAVeD (Current Dataset) | Global | Crash & Near Miss | text report & video | 2119 videos |

Currently, most automated vehicle crash data originate from official crash reports (*12*). While these reports serve an essential role in documenting significant collision events, they exhibit several inherent limitations that restrict the depth and comprehensiveness of safety assessment for ADAS-equipped vehicles. While video-based datasets like SAVeD provide rich, dynamic, and first-person insights into ADAS-equipped vehicle crashes and near-miss events, traditional textual crash reports remain the predominant data source in current research. However, these reports suffer from significant limitations that hinder comprehensive understanding and analysis of ADAS-equipped vehicle safety performance, highlighting the critical need for richer and more dynamic video data to better capture the complexities of ADAS driving scenarios.

First, traditional crash reports are typically textual, recording only structured information such as the time, location, vehicle type, and whether there were any injuries or fatalities. DMV collision and disengagement reports usually contain brief narrative descriptions relying on the subjective accounts of human drivers, lacking dynamic behavioral analysis of the crash process, such as detailed event sequence and real-time system responses. This results in a significant gap in understanding the full dynamics of the crash (*17*). Text-based crash reports often lack detailed behavioral data, specific perception failures, and crucial dynamic parameters like time-to-collision (TTC) during the incident (*18*). These omissions hinder in-depth analysis of crash causation and the optimization of autonomous driving systems. Another limitation is the lack of first-person perspective recordings, which makes it difficult to reconstruct the actual perception and control processes of the driver or vehicle during the event, an essential gap that first-person video data can effectively address. Compared to static and retrospective crash reports, crash and near-miss video datasets provide synchronized, dynamic, and context-rich information, enabling more accurate post-hoc analysis and system diagnosis (*19*).

Second, traditional crash reports focus primarily on actual crashes, paying insufficient attention to the vast number of near-miss incidents, limiting their usefulness in improving autonomous driving safety(*20*). These reports typically document only major crashes involving property damage or personal injury, while systematically overlooking near-miss events (*21*). One reason for the systematic neglect of near-miss incidents in official reports is that most regulatory frameworks only mandate documentation of crashes involving injury or property damage. As a result, a wide range of high-risk but non-crash scenarios remain invisible in formal datasets. Yet, near-miss incidents are highly informative—they not only reveal moments when ADAS performance approaches failure but also capture instances where the system successfully avoids collisions, navigates complex interactions with human-driven vehicles, and protects the automated vehicle itself (*22*). Such events provide invaluable insights into the system's real-world capabilities, limitations, and interaction strategies under challenging scenarios.

Third, not every minor crash results in a police report or gets recorded officially, leading to data gaps and incompleteness regarding the actual incidence and detailed characteristics of ADAS-equipped vehicle crashes in the real world. Fatal or severe crashes are more likely to be documented by authorities, while more than two-thirds of minor crashes and those causing only property damage go unreported (*23*).

More importantly, existing reports often lack deep insight into the ADAS driving system's own errors, particularly those that do not result in crashes or near misses (*18*). Examples include failures in perception or decision-making, such as misidentifying school buses or police vehicles, or incorrectly reading speed limit signs. These microscopic system faults may not directly cause crashes. However, systematic collection and analysis of such micro-level system failures, many detectable only through continuous video observation, are crucial for advancing ADAS safety research.

### Background and Motivation for Proposing the SAVeD Dataset

To address the current data gaps in ADAS-equipped Vehicle safety research and provide more detailed and comprehensive support for in-depth analysis of ADAS vehicle performance, we introduce SAVeD (Social media-based ADAS-equipped Vehicle event Dataset). SAVeD is currently the largest publicly available video dataset focusing on first-person perspective recordings of ADAS-equipped vehicle collisions and near-miss events. The dataset innovatively collects data from mainstream social media platforms such as YouTube, BiliBili, and Douyin, encompassing a wide range of real-world scenarios and diverse driving environments.

Compared to traditional structured crash reports, video data offers more vivid and continuous event records, richly illustrating vehicle dynamic behavior, real-time driver reactions, interactions with surrounding traffic participants, and feedback from ADAS-equipped driving system interfaces(*19*). These insights not only compensate for the limitations of traditional reports but also provide an intuitive perspective on the perception and decision-making processes during incidents. Through in-depth video analysis, we gain a more accurate understanding of how ADAS systems perform and where their limitations lie in complex real-world traffic conditions, identifying potential safety risks and providing a solid data foundation for system design, testing, and optimization.

Moreover, video data authentically captures the field of view of both the driver and the ADAS, revealing detailed perception and decision-making information that traditional textual reports cannot match. The establishment of SAVeD not only enriches the resources available for researching ADAS-equipped driving crashes and near-misses but also offers valuable empirical evidence for future improvements in ADAS safety performance. Currently, most existing first-person accident datasets lack detailed annotations. Specifically, video datasets such as DADA-2000 *(34)*, CCD *(35)*, and AV-TAU *(36)* primarily focus on training large language models (LLMs) through question–answer pairs, which results in relatively coarse labels compared to SAVeD. In contrast, SAVeD provides highly detailed annotations, making it a more valuable resource for traffic safety research. Furthermore, its richness can contribute to more effective LLM training by offering finer-grained contextual information.

### Overview of the SAVeD Dataset: Event Types and Annotation Approach

The SAVeD dataset is constructed to comprehensively cover various types of ADAS-equipped driving events, providing a rich event repository. These videos were collected from mainstream social media platforms such as YouTube, BiliBili, and Douyin, enabling large-scale access to first-person event footage that is rarely available in official datasets.

**Table 2** presents the description of this dataset with three main components: 1,040 meticulously annotated videos(614,236 frames) of ADAS-equipped vehicle collision events; Over 600 videos(306,781 frames) capturing near-miss incidents involving ADAS-equipped vehicles; 477 videos(214,650 frames) documenting system errors, such as perception or decision-making faults, SAVeD includes over 2,000 first-person driving videos that span a wide spectrum of real-world ADAS-equipped vehicle behavior, including collision events, near-miss incidents, and non-collision scenarios in which perception or decision-making errors were observed, providing a rich resource for comprehensive ADAS-equipped vehicle safety analysis.

**TABLE 2 SAVeD Dataset components and description**

| Category | Sample size | Description |
|---|---|---|
| ADAS-equipped vehicle collision events | 1,040 | First-person perspective crash video captured by an ADAS-equipped vehicle's onboard camera. |
| near-miss incidents involving ADAS-equipped vehicles | 602 | Near-miss event recorded by an ADAS-equipped vehicle-mounted camera; ADAS-equipped vehicle performed evasive action. |
| Non-collision or non–near-miss event involving system errors in the ADAS-equipped vehicle | 477 | The vehicle is initially controlled by the ADAS. Due to perception or decision-making errors, the system fails to respond appropriately, triggering a disengagement where the human driver manually takes over to correct the error. |
| Total | 2,119 | |

For the collision and near-miss videos, we performed highly detailed and in-depth manual annotations. Although recent advances in vision-language models (VLMs) have significantly improved video understanding(*24*), due to the complexity of ADAS-equipped vehicles driving events and the high accuracy demands on detail, we relied heavily on high-precision manual labeling(*25*). These annotations encompass 27 distinct dimensions, covering temporal and spatial information, vehicle status, environmental conditions, event progression, system behavior, and human interventions, ensuring both depth and breadth of data to support advanced analyses.

Notably, SAVeD offers unique value in capturing and analyzing near-miss events and system errors. Traditional crash reports typically record only collisions with severe consequences, while numerous near-miss incidents—where the ADAS successfully intervened to avoid crashes despite human driver limitations such as fatigue, reckless behavior, or limited visibility—remain largely undocumented(*26*). These "almost-crashes" provide invaluable insights into the safety boundaries, response capabilities, and decision robustness of ADAS in complex real-world scenarios. By analyzing these near-miss cases, we can identify conditions under which the system effectively mitigates risk and where potential vulnerabilities remain.

Similarly, collecting videos of system errors that were corrected or annotated by human supervisors allows us to investigate perception flaws (e.g., misclassification or missed detection) and suboptimal decision-making (e.g., path or speed planning errors) without the cost of actual crashes. These seemingly minor errors are critical learning points for future improvements in ADAS-equipped vehicles driving safety and reliability.

The SAVeD dataset primarily focuses on ADAS-equipped vehicles, which broadly include those that allow driver hands-off under supervision and those that require continuous driver engagement.

**Methods**

*Social Media Video Acquisition Process*

We developed a Python script to automatically search over 300,000 videos using keywords like "ADAS crash" and "self-driving vehicle near miss." All videos were manually reviewed and trimmed to retain only complete sequences of crashes, near-miss events, and ADAS-related errors. Sourced primarily from YouTube, low-quality or irrelevant clips were excluded to ensure high data quality for ADAS behavior and crash analysis research.

***Structured Annotation Framework for ADAS-equipped vehicle Incident Videos***

To support efficient and standardized labeling of ADAS-related incident videos, we developed a lightweight, Python-based annotation tool with a terminal-based interface. Annotators can input key attributes, including environmental conditions, traffic context, ADAS status, vehicle maneuvers, and critical timing (e.g., avoidance start, human takeover). The tool emphasizes temporal labeling, such as pinpointing crash frames and identifying reasons for human intervention. It supports keyboard navigation, revision, auto-fill for repeated inputs, and structured CSV output. The tool and sample data are shared with the dataset release to facilitate broader research use.

As shown in **Figure 1**, the SAVeD dataset was constructed through a multi-stage pipeline starting from video retrieval using Python keyword scripts, followed by human selection, frame extraction, and both macro- and micro-level labeling using Python-based annotation tools.

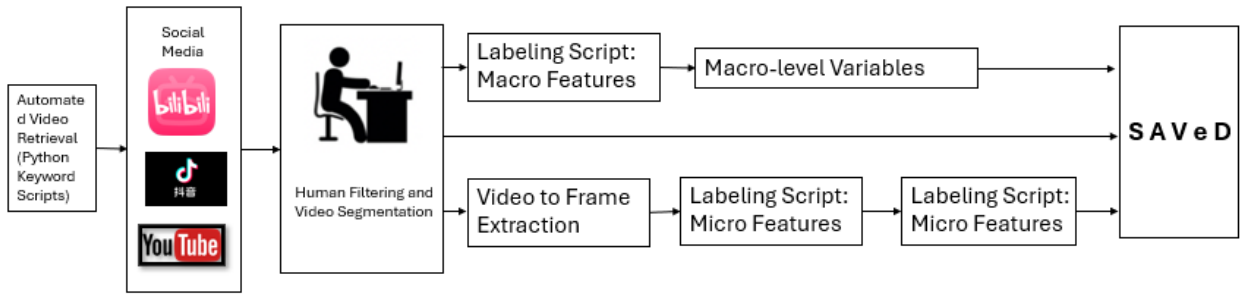

**Figure 1. Pipeline for SAVeD dataset creation from video retrieval to structured annotation**

***Variables extracted in SAVeD***

We conducted detailed video-based annotation of ADAS-involved crashes and near-miss events, extracting a wide range of behavioral and environmental variables. For crash cases, annotations covered lighting, weather, and road surface conditions; crash details such as impact type, involved objects, and number of vehicles; vehicle-level attributes including model, ADAS usage, and primary damage location; as well as roadway context (e.g., road type, gradient, lane count, traffic flow). Outcome-related variables, such as fault attribution, injury presence, and estimated repair cost, were also recorded.

**Table 3** summarizes the variable names, descriptions, and possible values used in this study. Detailed definitions and annotation examples are available on the project's GitHub repository to facilitate reproducibility and reuse.

**TABLE 3 summaries of the variables' description**

| Variable | Possible Values | Description and note |
|---|---|---|
| Lighting condition | Dark – Lighted, Dark - Not Lighted, Daylight | lighting condition of the scene during the crash or near-miss |
| Weather condition | Clear, Cloudy, Fog/Smog, Rain, Snow | weather conditions during the crash |
| Road surface | Dry, Wet, Snow/Ice | Indicates the road surface state |
| Fault attribution | Yes, No, Both at Fault | Whether the ADAS-equipped vehicle was considered at fault |
| Impact configuration | Front to Front, Front to Rear, Rear to Rear, Front to Side, Rear to Side, Sideswipe (Opposite Direction), and Sideswipe (Same Direction), other | Crash configuration or impact type. |
| Total vehicles involved | 1, 2, 3, 4, 5, 6, over 6 | Total number of vehicles involved in the crash. |
| Object collided with | Pedestrian, Vehicle, Bicycle, Infrastructure, Animal, other | Type of object that the AV collided with. |
| ADAS-equipped vehicle type | Car, Truck, SUV, other | type of the ADAS-equipped vehicle involved. |
| Damaged area on ADAS | Front Center Bumper, Front Left Bumper, Front Right Bumper, Left Front Door, Left Front Fender, Left Rear Door, Left Rear Fender, Rear Center Bumper, Rear Left Bumper, Rear Right Bumper, Right Front Door, Right Front Fender, Right Rear Door, Right Rear Fender, Roof, Undercarriage, Windshield, Rollover, Overturn, other | Location of the most significant damage on the ADAS-equipped vehicle. |
| Damaged area on the other vehicle | (Same as above) | Most damaged part of another vehicle |
| ADAS status | hands-off under supervision, those that require continuous driver engagement. | Indicates the ADAS (Advanced Driver Assistance System) status at the time of crash. Determined via driver behavior in video (e.g., hands on wheel, gaze direction), subtitle cues (e.g., "engaged FSD"), and vehicle model/year. Only videos with clearly identifiable ADAS status are included. |
| Road type | Highway/Freeway, Local city, Signalized intersection, non-signalize intersection, Parking lot, Traffic circle, Rural Road, Unpaved Road, Bridge, Tunnel, Work zone, other | the type of road or traffic environment |
| Crash type | Angle, Head On, Left Turn, Right Turn, Sideswipe, Off Road, Rear End, other | Categorizes the crash scenario |

| Variable | Possible Values | Description and note |
|---|---|---|
| Road grade | Yes (Flat), Up (Uphill), Down (Downhill) | Indicates whether the road is flat, uphill, or downhill |
| Injury outcome | No, Yes (not bad), Yes (bad), Dead, idk | Severity of personal injury resulting from the crash (from subtitles). |
| Cause of crash | Failure to Yield, Following Too Closely, Improper Lane Change, Running Red Light, Running Stop Sign, Backing Without Caution, Wrong-way Driving, Slippery Road (Rain, Snow, Ice), Animal on Road, other | Primary cause contributing to the crash. |
| Type of opposing vehicle | Big vehicle (Bus/Truck), Middle vehicle (Pickup/SUV), Small car, Bike or Pedestrian, other | Another type of vehicle crashed into. |
| ADAS vehicle's movement pre-crash | Stopped, Proceeding Straight, Making Right Turn, Making Left Turn, Backing, Parked, other | ADAS-equipped vehicle behavior immediately before the crash. |
| Other vehicle's movement pre-crash | (Same as above) | Another vehicle behavior immediately before the crash. |
| ADAS vehicle's action to avoid crash | Right turn, Left turn, Decelerating, Accelerating, No Action, other | ADAS-equipped vehicle action to avoid the crash |
| Another car's action | Right turn, Left turn, Decelerating, Accelerating, No Action, other | Another vehicle action to avoid the crash |
| Time AV attempted to avoid crash | (Same as above) | Action taken by ADAS-equipped vehicle to avoid the crash. |
| Time another car avoided | (Free text, numeric) | Action taken by the other vehicle to avoid the crash. |
| Crash location | (Free text, numeric) | Display state name for U.S. crashes; otherwise, show country name. |
| Number of lanes (one direction) | 1 lane, 2 lanes, 3 lanes, 4 lanes, 5 lanes, over 5 lanes | The number of lanes in one direction. |
| Traffic flow | Congested, Red Light Stopped, Moderate Traffic, Light Traffic, other | Summarizes traffic flow at the time of crash. Light Traffic means few or no vehicles present; Congested indicates heavy traffic with little movement; Red Light Stopped means vehicles stopped at a red light; all other cases are classified as Moderate Traffic. |
| Estimated repair cost | (Free text, dollar amount) | Some users provide repair/insurance receipts or describe costs in subtitles. If "total loss" is mentioned, the value is recorded as total |

**Figure 2(a) and Figure 2(b)** illustrate the pre-crash scene. Based on video footage and on-screen captions, the environment featured daylight, clear weather, dry road surfaces, and light traffic. The crash occurred at a flat, unsignalized, single-lane intersection. The opposing vehicle - a small car - was traveling in the wrong direction and made a right turn, posing a head-on threat. The ADAS-equipped vehicle was initially proceeding straight and executed an evasive right turn to avoid the collision. According to the captions and vehicle model, the ADAS system was engaged but required continuous driver supervision.

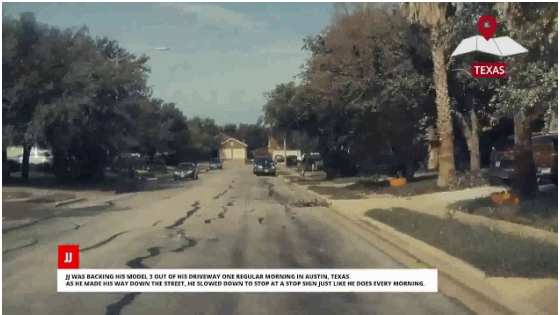

**(a) Pre-Crash Environment Overview**

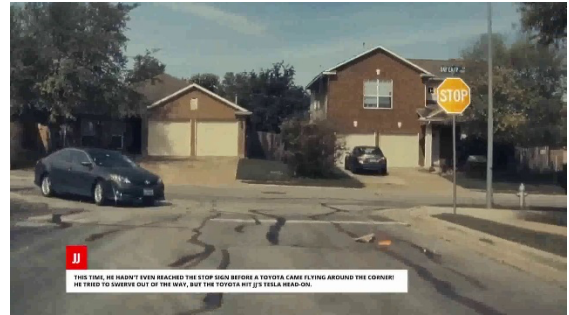

**(b) Onset of Pre-Crash Maneuver**

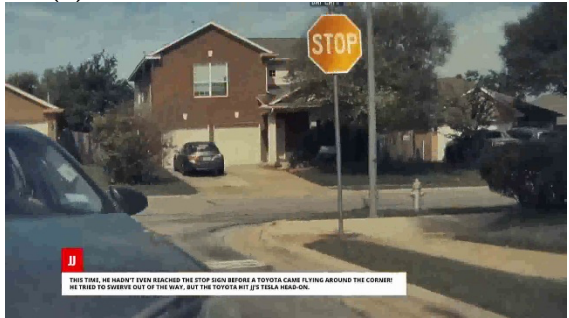

**(c) Crash Moment**

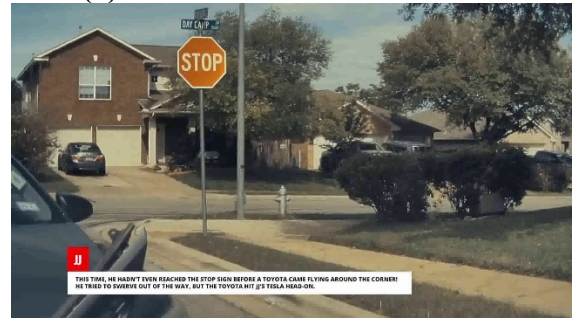

**(d) Post-Crash Outcome**

**Figure 2. Annotated Example of an ADAS-equipped vehicle-Involved Crash Case**

**Figure 2(c) and Figure 2(d)** depict the crash moment and post-crash outcomes. By analyzing frame-by-frame video data, it was determined that the ADAS-equipped vehicle initiated the evasive maneuver 1.2 seconds before the collision. The crash type was head-on, with both vehicles sustaining damage to their front-center sections. Two vehicles were involved, and the primary cause was identified as wrong-way driving. According to the uploader's description, a passenger in the ADAS-equipped vehicle sustained serious injuries and was hospitalized. Based on available visual and contextual evidence, fault was attributed to the opposing vehicle. No information regarding insurance compensation was provided in the video.

**Dataset Record**

The SAVeD dataset consists of three categories of ADAS-equipped vehicle events, as summarized in **Table 2**:
- 1,040 collision videos (614,236 frames),
- over 600 near-miss videos (306,781 frames), and
- 477 system-error videos (214,650 frames),

For each event, a corresponding CSV record file is provided. The CSV files include:
- the source URL of the video,
- the start time and end time of the annotated event segment,
- and all frame-level variables listed in **Table 3**.

To ensure compliance with YouTube and other social-media platform policies, original video files are not redistributed. Users must obtain the videos directly from their source platforms using tools such as youtube-dl (*49*), following all applicable terms of service.

All data entries follow a consistent naming scheme: AV_crash.csv, AV_nearmiss.csv. The full set of CSV files, together with the complete URL list and annotation metadata, will be made available in the accompanying GitHub repository upon publication.

**Dataset Composition and Visualization**

*Risk Avoidance of ADAS*

In the SAVeD dataset, the variable "ADAS-equipped vehicls_avoid_seconds_before_crash" is used to capture the time (in seconds) before the crash at which the ADAS-equipped vehicle initiated an avoidance maneuver, with the distribution shown in **Figure 3**. This is a highly valuable indicator of system responsiveness in real-world scenarios.

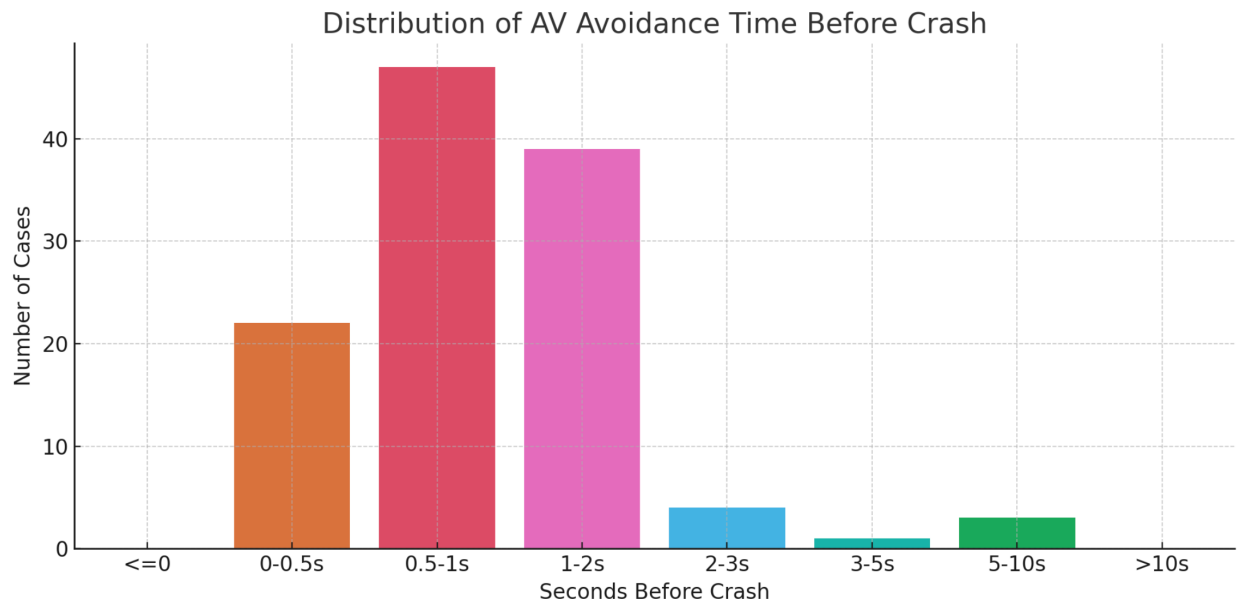

**Figure 3. Distribution of ADAS-equipped vehicle avoidance time before crash in SAVeD dataset.**

Based on real-world video annotations, ADAS-equipped vehicles in our dataset initiated evasive actions on average 1.07 seconds before collision, offering high-resolution temporal insights rarely found in existing datasets. Although existing simulation and naturalistic studies report human driver reaction times typically ranging from 0.8–1.3 seconds for braking and 0.6–1.2 seconds for steering (*28, 29*), these estimates are primarily derived from controlled experiments or simulated driving scenarios, rather than real-world crash footage. In contrast, our dataset focuses on actual ADAS-equipped vehicle-involved crashes captured through publicly available videos, allowing for empirical measurement of evasive maneuvers under real crash conditions. This distinction is crucial, as simulated environments may not fully capture the stress, unpredictability, and timing dynamics inherent in real-world emergencies. While this allows for empirical examination of ADAS-equipped vehicle response timing under real crash conditions, a direct performance comparison with human drivers is not yet available.

To address this, future extensions of the dataset will incorporate human-driven vehicle crash cases with comparable temporal annotations, enabling robust side-by-side comparisons of avoidance behavior. This planned expansion will support a more systematic evaluation of ADAS-equipped vehicle decision-making relative to human drivers, advancing both behavioral modeling and safety benchmarking efforts in ADAS-equipped vehicle research.

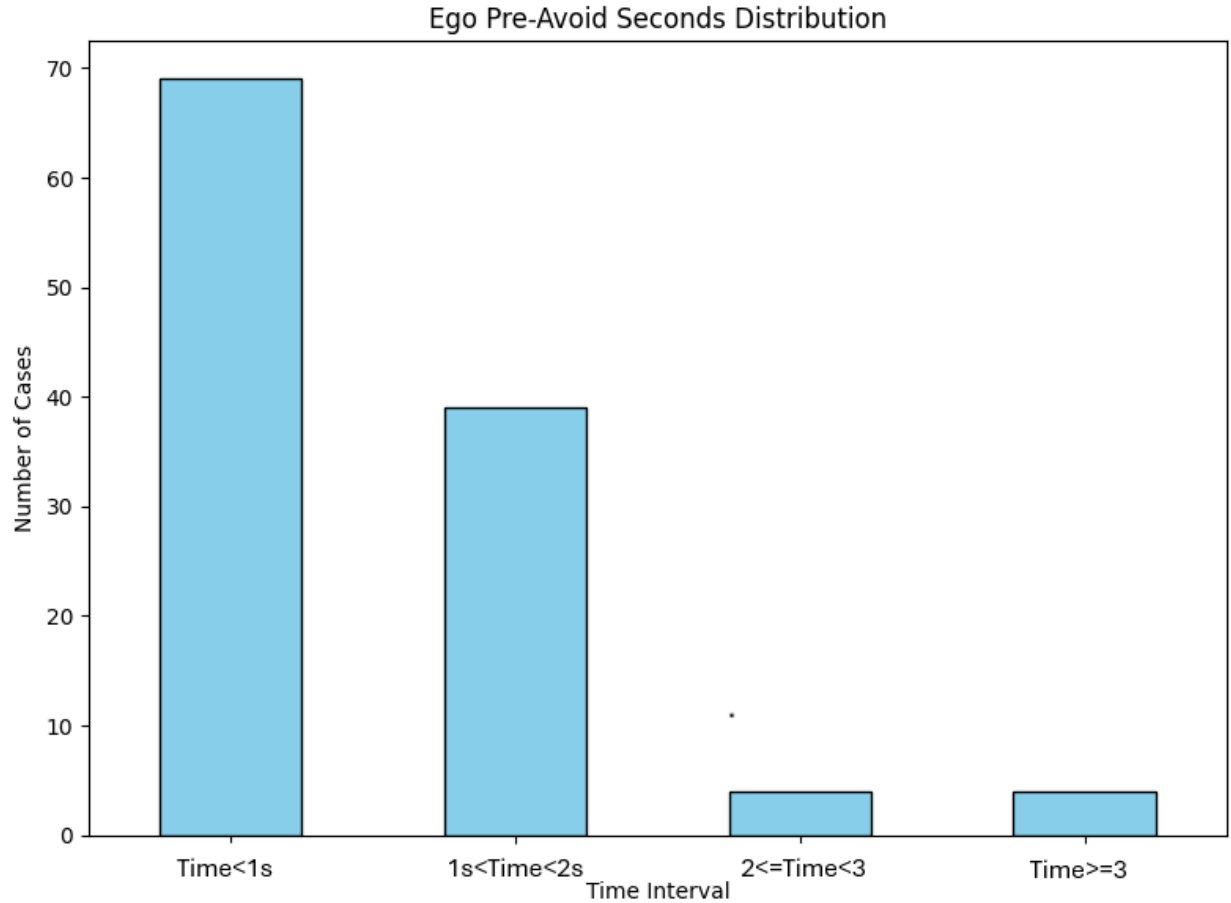

**Figure 4. Distribution of ADAS-equipped vehicle avoidance time before crash in near-miss events in SAVeD**

In addition to real-world crash events, key variables were also extracted from a curated set of near-miss incidents involving ADAS-equipped vehicles. One notable variable is the AV avoidance time before a potential collision, capturing the time interval between the initiation of an evasive maneuver and the moment a crash would have occurred. This information, annotated frame-by-frame from video footage, offers high-resolution temporal insights into ADAS-equipped vehicle behavior under imminent risk. As shown in the distribution plot **Figure (4)**, most ADAS-equipped vehicles responded within 1–3 seconds before a near-miss, which is consistent with but slightly faster than human drivers' reaction times reported in simulation studies. The inclusion of near-miss cases not only strengthens the behavioral granularity of our dataset but also sets a critical foundation for future comparisons with human-driven vehicle reactions and proactive risk intervention modeling.

### *ADAS Errors: Categorization and Implications*

The third component of the dataset focuses on various types of ADAS-equipped vehicle system errors encountered during driving. Although these errors do not necessarily result in actual crashes or near-miss events, they provide valuable insights for evaluating the safety performance of ADAS-equipped vehicle systems. Based on video analysis, these errors include, but are not limited to, failure to recognize traffic signs, merging errors, pedestrian neglect, overlooking speed limits, lane departures or off-road driving, running red lights, misinterpretations caused by snow-covered conditions, unexplained sudden braking (commonly referred to as "phantom braking"), failure to yield to emergency or school vehicles, and missed detection of school zones or other traffic control signals.

Such diverse and representative error cases offer rich data for advancing perception modules, decision-making algorithms, and complex scenario modeling under dynamic traffic conditions. When combined with other multimodal datasets, such as radar point cloud dataset (*29*), LiDAR-based datasets (*3, 31*), visual perception datasets(*1*) and depth estimation datasets (*32, 33*), SAVeD significantly contributes to improving ADAS-equipped vehicle systems' responsiveness to challenging environments and atypical traffic signals, thereby enhancing the overall safety and robustness of ADAS driving technologies. **Figure 5** presents the number of collected video examples corresponding to each error category in the SAVeD dataset. Rather than indicating real-world occurrence frequency, this distribution reflects the relative scale and coverage of each error type within the dataset, providing researchers with a useful overview of its content structure and focus areas.

All included videos feature vehicles with visible ADAS engagement operating in a "hands-off under supervision" mode. Observable system behaviors, such as perception or decision-making failures (e.g., not slowing in school zones), are corrected by human intervention. Videos with unclear driver actions or ambiguous ADAS status were excluded to ensure consistency and reliability.

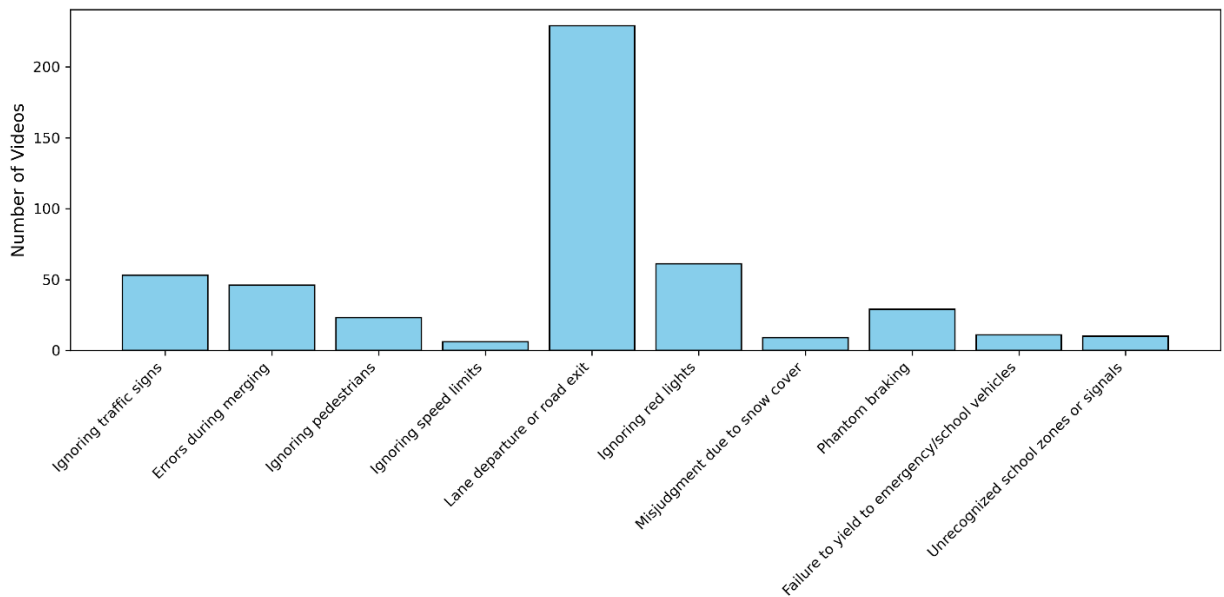

**Figure 5. Number of Collected Videos per ADAS-equipped Vehicle Error Type in the SAVeD Dataset**

### *Crash Characteristics and Repair Impact Analysis*

**Figure 6(a)** shows average repair costs by damaged area in ADAS-equipped vehicle crashes. The highest costs are linked to roof, undercarriage, front bumper, and doors—especially rollovers or roof impacts, which can exceed $38,000 due to structural damage and embedded sensors. While some cases lack cost estimates, trends highlight high-impact zones for design and insurance considerations.

**Figure 6(b)** presents crash type distribution. Rear-end and sideswipe collisions are most common, pointing to limitations in time-to-collision estimation, braking response, and blind-spot monitoring—areas where ADAS perception and planning need improvement.

**Figure 6(c)** lists top crash causes, led by "Following Too Closely" and "Improper Lane Change," followed by "Failure to Yield" and "Running a Red Light." These behaviors reflect challenges ADAS systems face in dynamic traffic and right-of-way negotiation.

**Figure 6(d)** shows crash distribution across road types. Most crashes occur on highways/freeways and local city roads, with many at intersections. This suggests ADAS-equipped vehicles struggle with high-speed merging and complex urban scenarios, reinforcing the need for improved system robustness in varied environments.

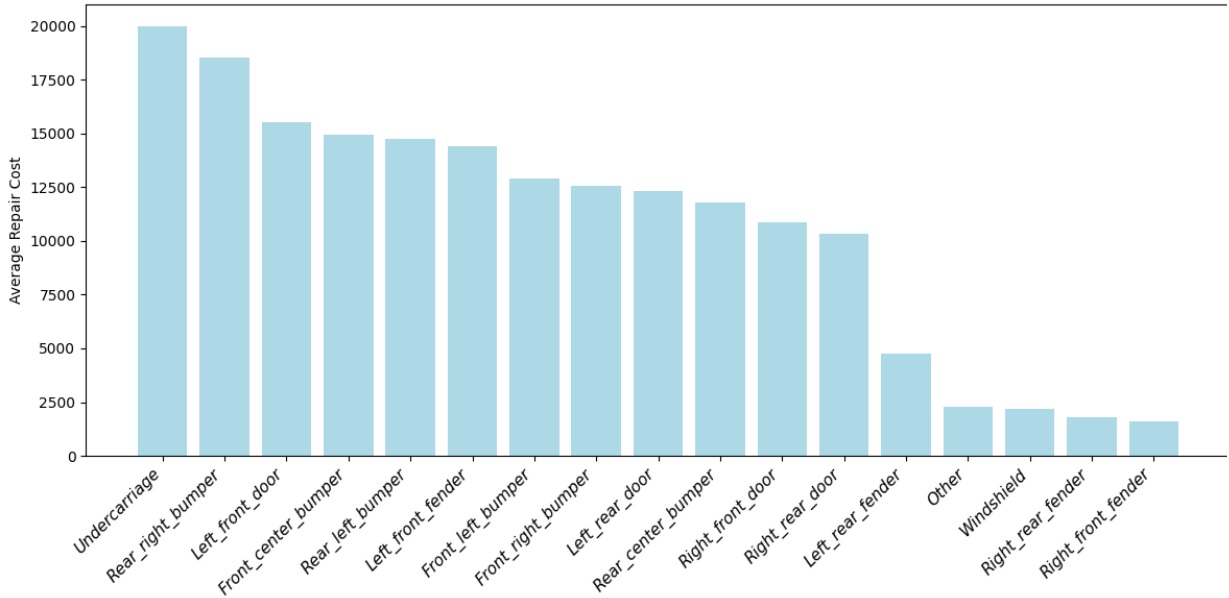

**(a) Average Repair Cost by Damaged Area**

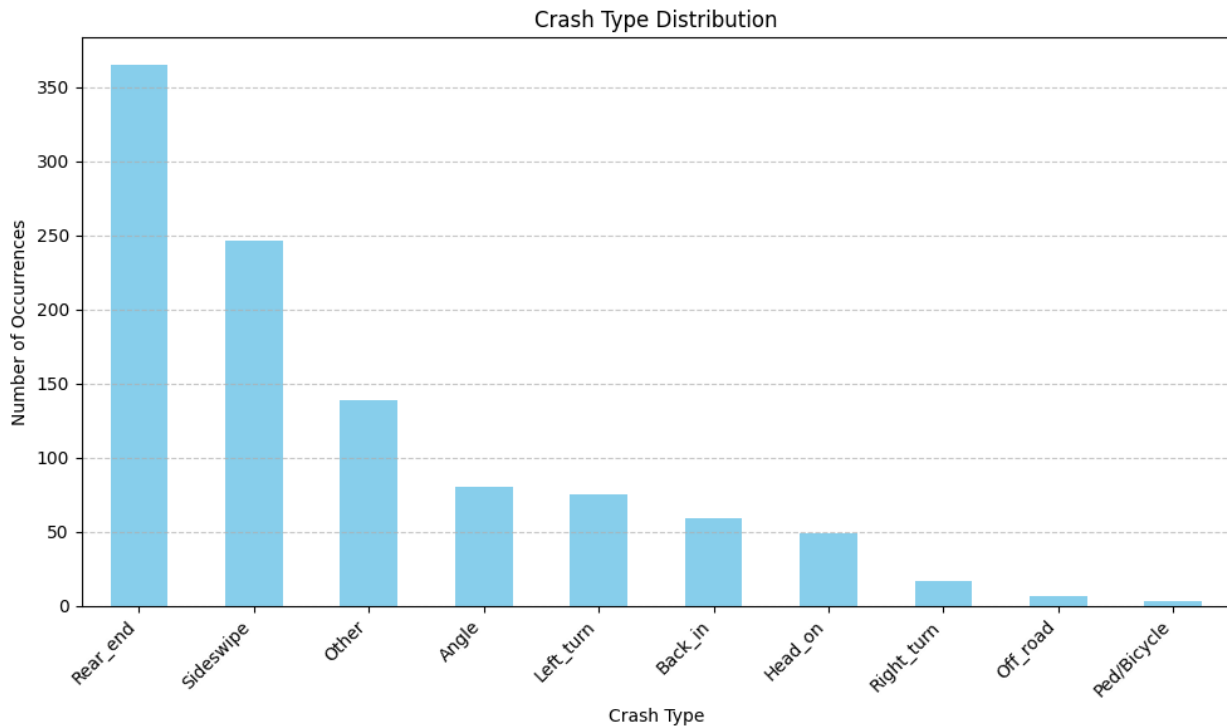

**(b) Crash Type Distribution**

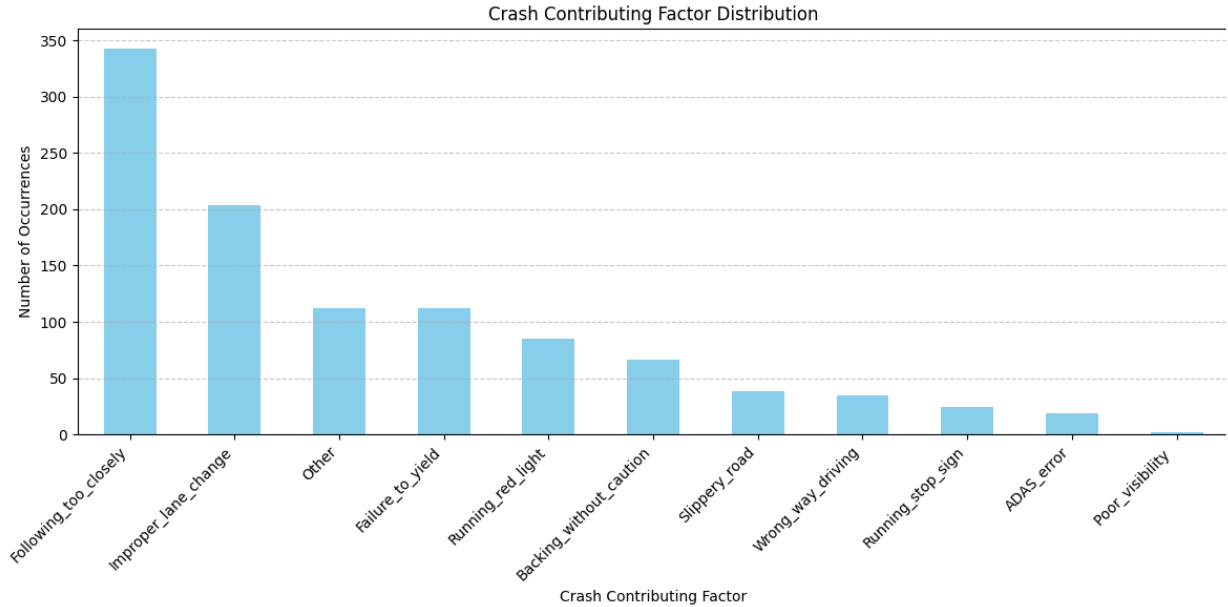

**(c) Crash Contributing Factor Distribution**

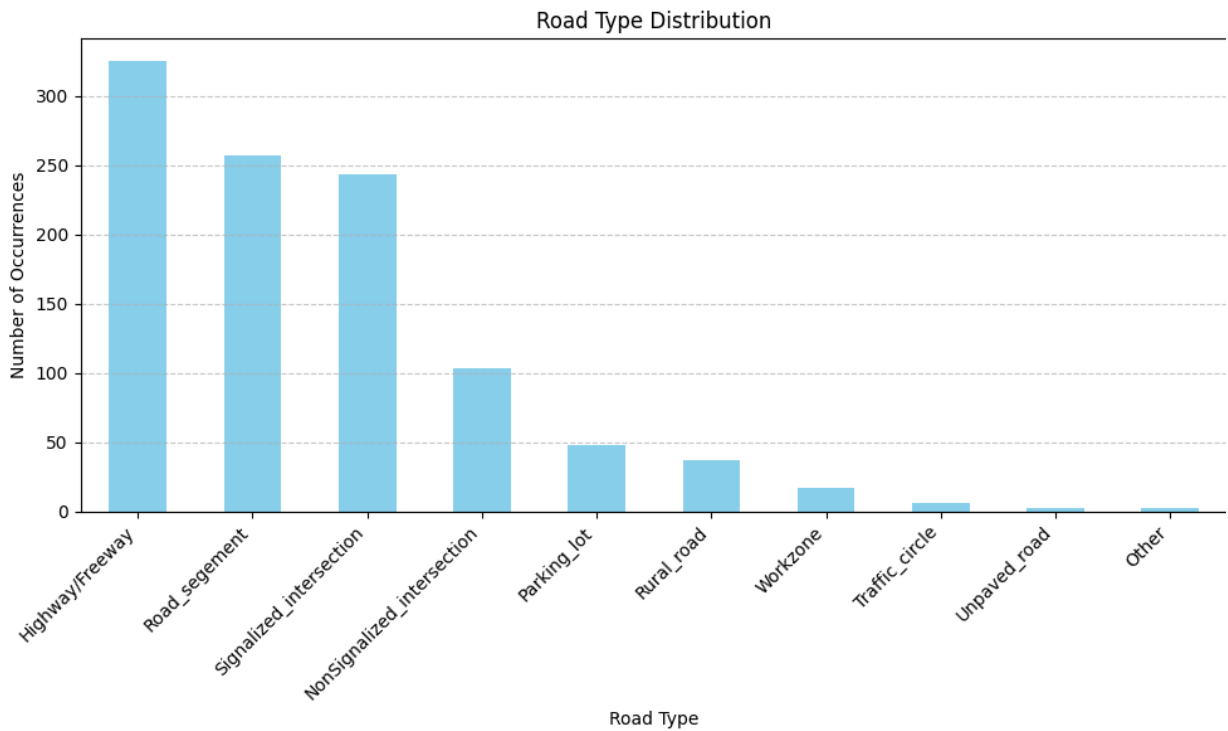

**(d) Road Type Distribution**
**Figure 6. Crash Characteristics and Risk Factors in ADAS-equipped vehicle-Involved Incidents**

### *Macro-Level Environmental Annotations and their Role in Understanding Crash Contexts*

The SAVeD dataset provides rich macro-level annotations of the driving environment, including weather conditions (**Figure 7(a)**), road surface status (**Figure 7(c)**), and lighting conditions (**Figure 7(e)**). These annotations allow for in-depth analysis of external factors contributing to crashes. Notably, the majority of incidents occurred under clear weather, dry road surfaces, and daylight conditions, which highlights the fact that many crashes take place even in seemingly safe environments.

**Figure 7** presents the distribution of ADAS-equipped vehicle fault attribution under varying environmental conditions, including weather (**Figure 7(b)**), road surface (**Figure 7(d)**), and lighting (**Figure 7(f)**). The results consistently show that ADAS-equipped vehicles are more likely to be held responsible for crashes in adverse conditions—such as rain, snow, wet or icy roads, and dark unlit environments. These factors may degrade sensor performance or impact system decision-making. Notably, the combination of rainy weather, slippery roads, and unlit darkness corresponds to the highest fault attribution. In contrast, fault rates are significantly lower under clear weather, dry roads, and daylight, highlighting the performance gap between ideal and challenging scenarios. These findings underscore the need to improve ADAS robustness in complex environments to enhance safety and reliability.

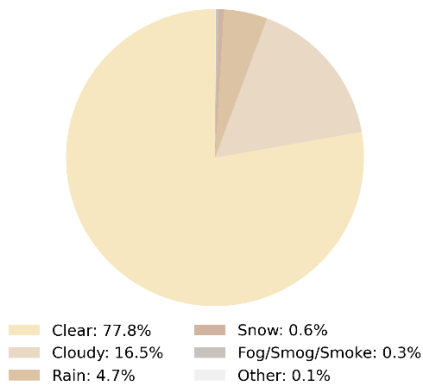

**(a) Weather Condition Distribution**

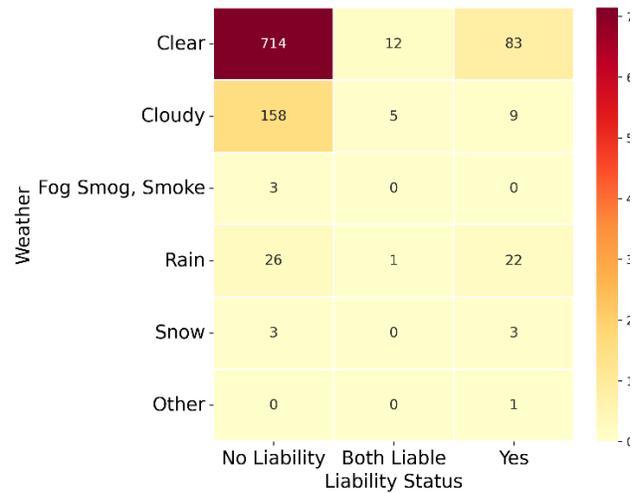

**(b) ADAS-equipped vehicle Fault vs Weather Conditions**

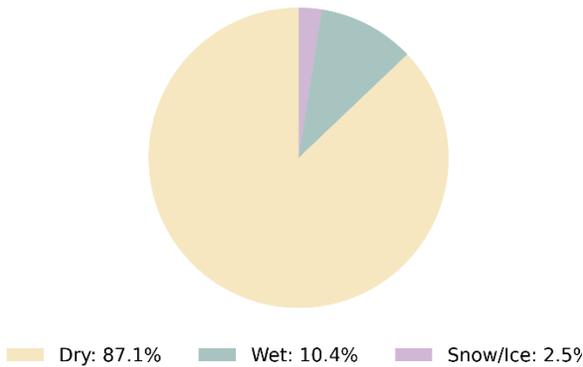

**(c) Road Surface Condition**

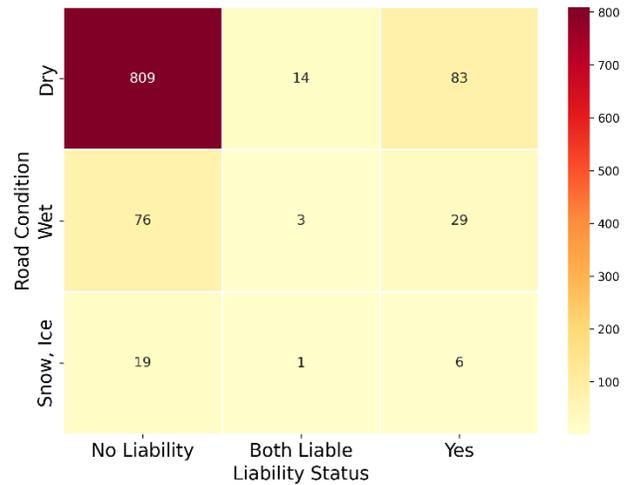

**(d) ADAS-equipped vehicle Fault vs Road Surface Conditions**

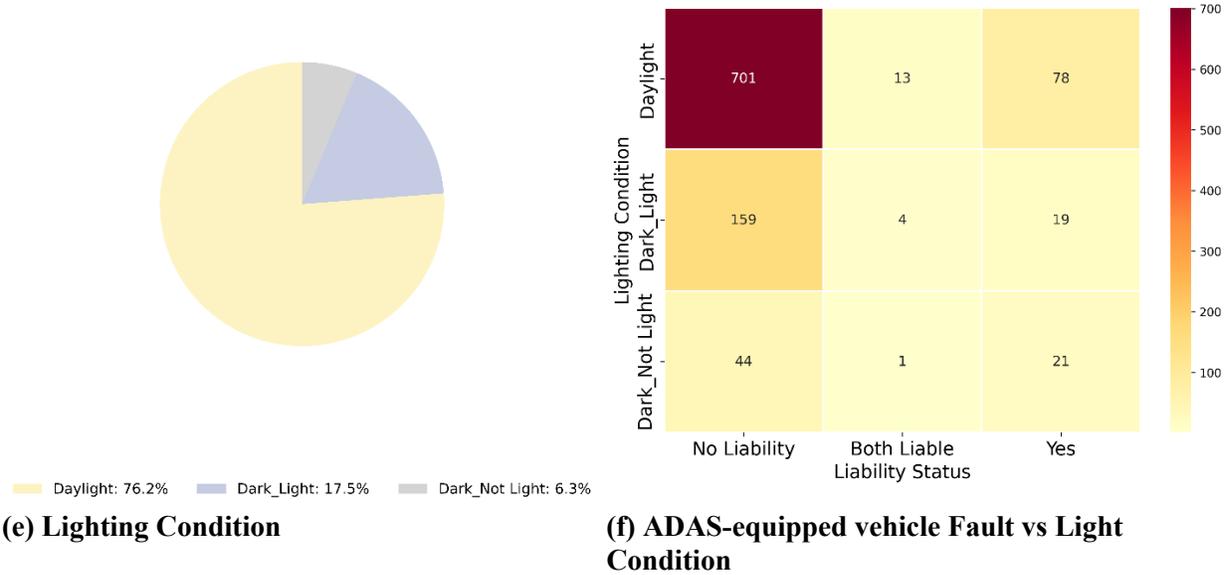

**(e) Lighting Condition**
Daylight: 76.2%  Dark_Light: 17.5%  Dark_Not Light: 6.3%

**(f) ADAS-equipped vehicle Fault vs Light Condition**

**Figure 7. Weather Condition, Road Surface and Lighting Condition**

As shown in the figure (**Figure 8**), in structured yet risk-prone environments such as highways, parking lots, and work zones, the hands-off under supervision mode exhibits a significantly higher crash proportion than the continuous driver engagement mode, by 12.9%, 7.78%, and 2.32%, respectively. All driving modes in this dataset have been carefully verified from the accompanying subtitles. This suggests that current systems may struggle with perception or decision-making in these contexts, indicating areas that require further improvement.

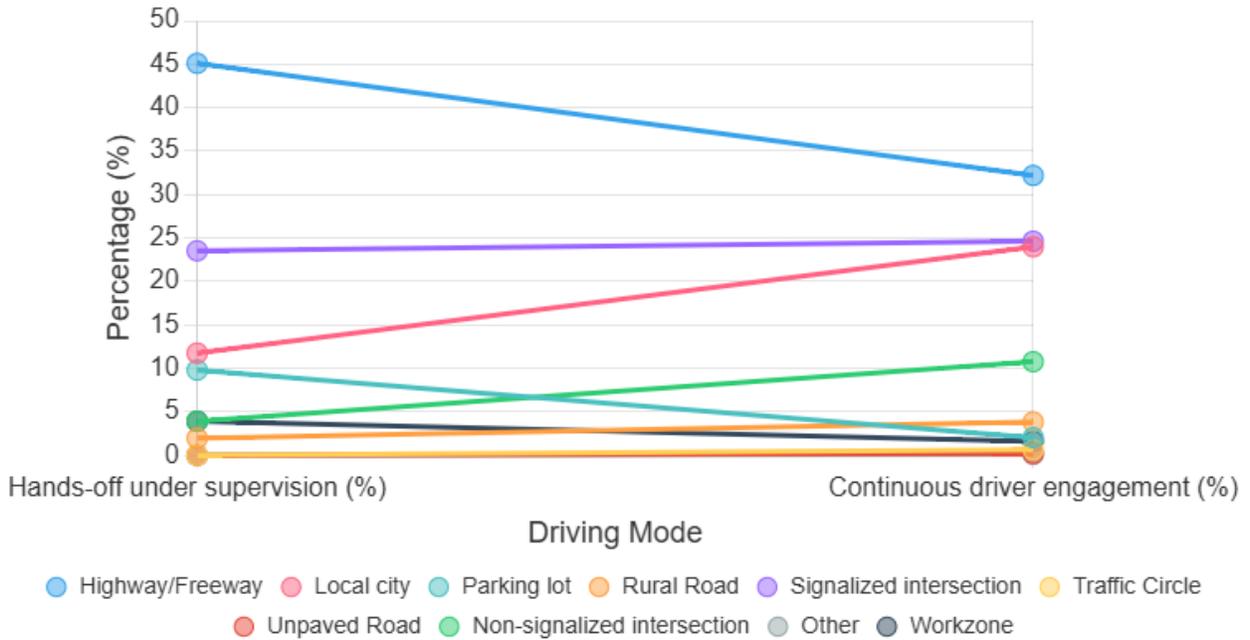

**Figure 8. Slope Chart of Hands-Off vs. Continuous Driver Engagement by Road Type**

In contrast, in more complex environments such as local streets, unsignalized intersections, and rural roads, the hands-off mode shows a lower crash share, suggesting better stability and adaptability when navigating dynamic and diverse traffic conditions.

**Figure (9)**: The hands-off mode's crashes are more concentrated in rear-end and sideswipe collisions, and the higher proportion of these crash types further reflects the ADAS system's cautious and defensive driving behavior in relatively low-speed, frequent traffic scenarios.

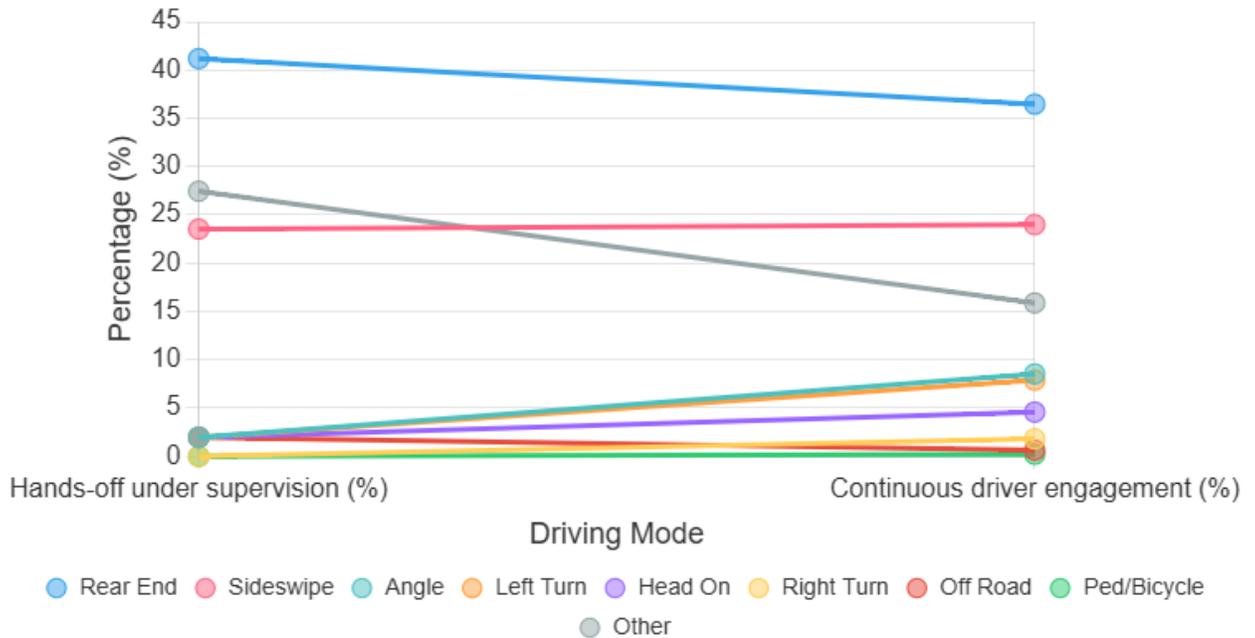

**Figure 9. Slope Chart of Accident Type Percentages by Driving Mode**

Meanwhile, the hands-off mode exhibits lower crash proportions in complex interactive scenarios such as angle collisions and left-turn crashes, demonstrating better adaptability. In contrast, the Autopilot mode shows a wider distribution across various crash types, possibly reflecting a greater diversity of risks and complexities under continuous human supervision.

### *Global Coverage and Cross-Cultural Potential*

The SAVeD dataset captures ADAS-equipped vehicle crashes and near-miss events worldwide, primarily involving well-known brands from China and the United States. Incident locations are visualized on maps, with U.S. events categorized by state and international events by country. This wide geographic coverage enables comparative analyses of ADAS performance and risks across different traffic cultures, infrastructure, and regulations—including distinctions such as left- vs. right-hand traffic (*31*).

**Figure 10 (a)** illustrates the global reach of the SAVeD dataset, highlighting countries (in red) where ADAS-equipped vehicle-related incidents were collected. The dataset spans over 40 countries, with the top ten countries annotated alongside their corresponding event counts.

**Figure 10 (b)** depicts the distribution within the United States, covering 45 states, including Hawaii and Alaska (not shown on the map). States where incidents occur are marked in red, and the top ten states are labeled with their respective event frequencies.

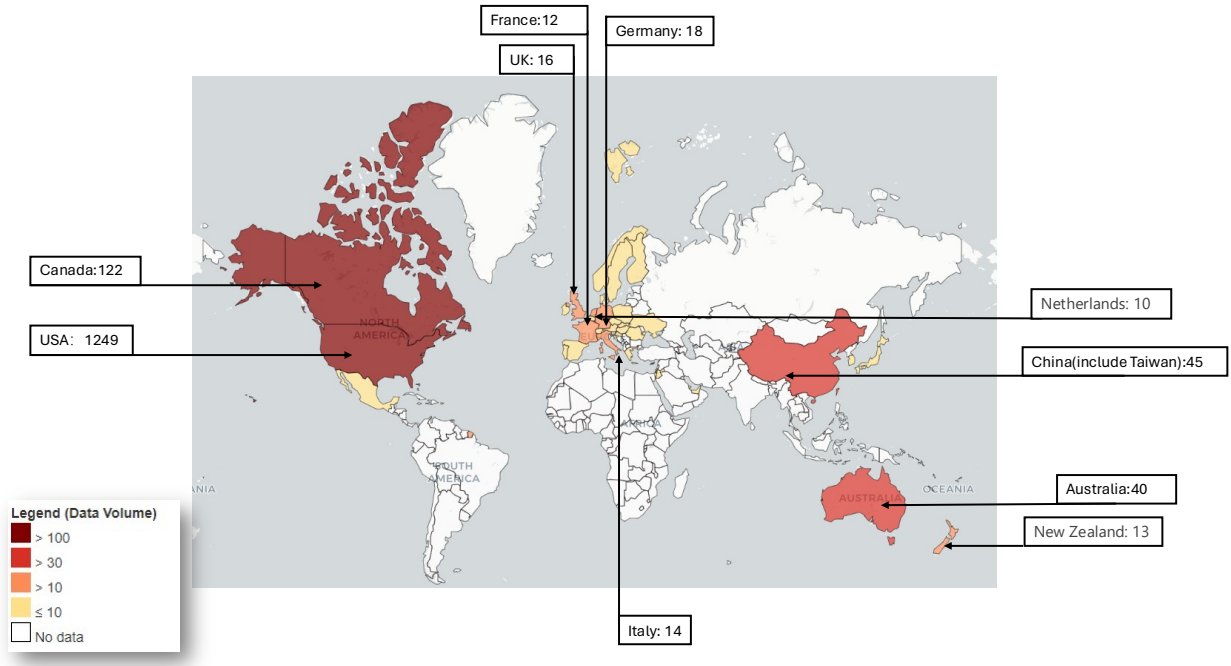

**(a) Events location around the world**

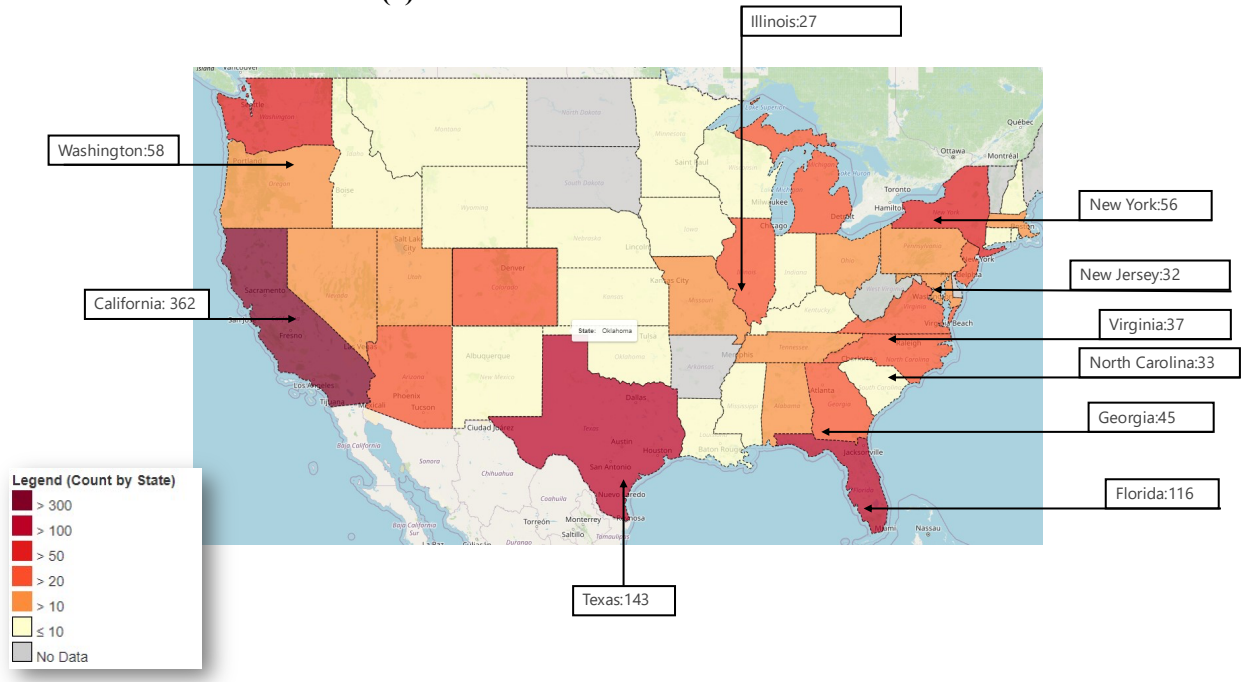

**(b) Events location around the USA**
**Figure 10. Event location**

### *Near-miss events*

The SAVeD dataset offers detailed annotations beyond basic crash data, covering environmental factors (e.g., weather, road surface), traffic conditions (lane configuration, flow), and behavioral responses of ADAS-equipped and surrounding vehicles. These enable in-depth analysis of vehicle interactions and near-miss dynamics for better risk assessment and safety system development.

**Figure 11** shows that both non-hands-free and supervised hands-free ADAS modes are widely used in avoidance maneuvers, with supervised hands-free systems used more frequently. Under "dark – lighted" conditions, supervised hands-free ADAS demonstrate higher usage and stronger avoidance performance, especially in low-light scenarios like night or dusk, indicating superior perception and decision-making.

Non-hands-free systems perform more reliably during daytime but decline under poor lighting, suggesting supervised hands-free systems have advantages in adverse conditions. Although supervised hands-free usage is currently low due to recent introduction, their effectiveness in high-risk environments indicates strong potential for future ADAS development.

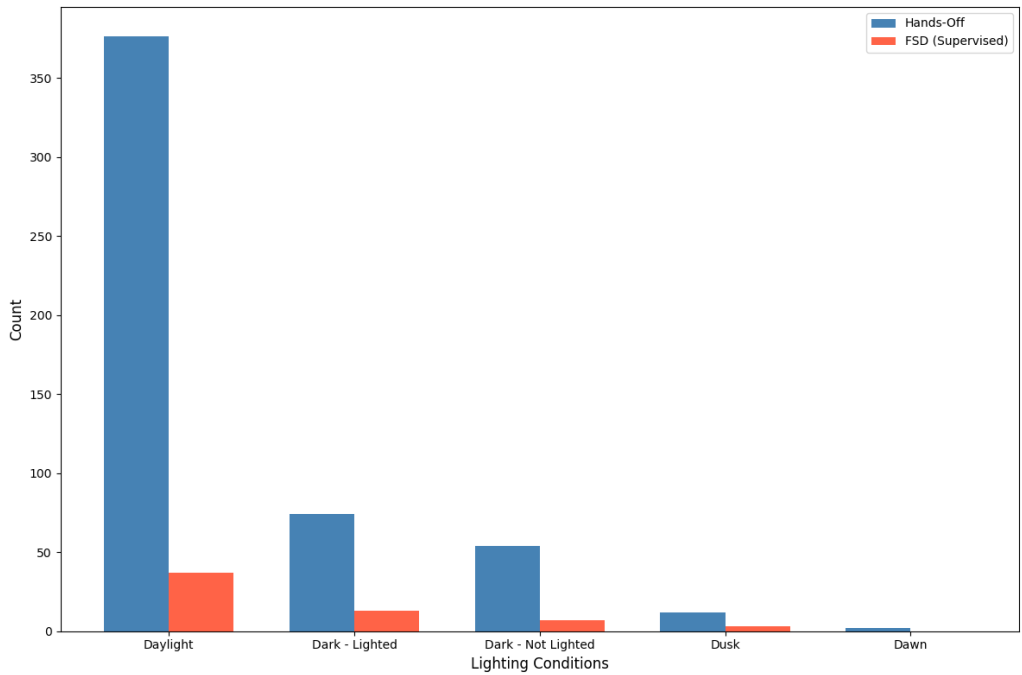

**Figure 11. crash-Avoidance Rates by Driving Mode and Light Condition**

### Vision-Based crash Reconstruction

This study proposes a vision-only pipeline for estimating Time-To-Collision (TTC) using monocular first-person-view (FPV) driving videos. The method relies solely on semantic segmentation and monocular depth estimation, without requiring GPS, IMU, LiDAR, or external calibration data. By leveraging semantic information and depth estimates from video frames, the method estimates dynamic object distances and velocities to compute TTC. The pipeline is illustrated in **Figure (12)**:

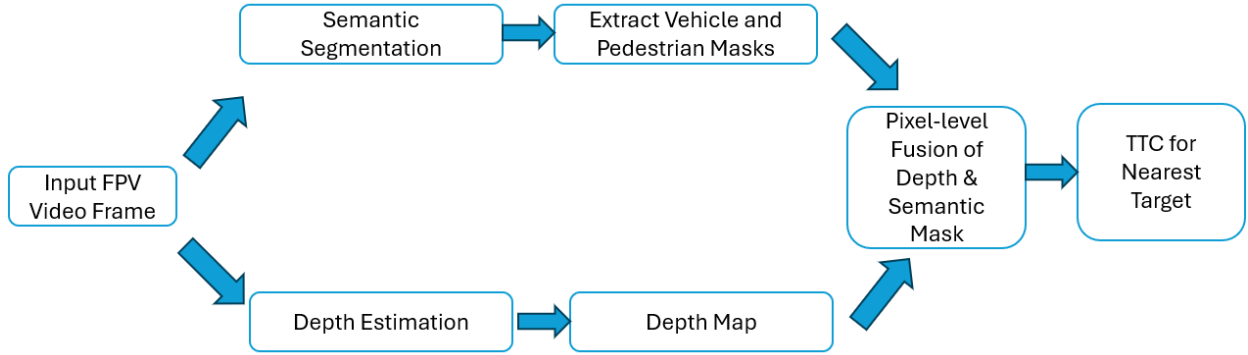

**Figure 12. Monocular Video TTC Estimation Pipeline**

To identify dynamic traffic participants in each video frame, we utilize a semantic segmentation model trained on the Cityscapes dataset (*1*)), implemented using the open-source MMSegmentation (*32*) framework, which has been widely used in existing studies (*33–35*). The model performs pixel-level classification on the input frame I, producing a semantic label map M(**Figure 13 (a)(c)**), where each pixel is assigned a class ID.

For each target object class (e.g., "vehicle"), the corresponding pixel region is extracted as:

$$R = \{(x, y) | M(x, y) = c\} \tag{1}$$

**Equation 1:** The region R consists of all pixels labeled as class c in the semantic segmentation map. Denotes the semantic class label of interest (e.g., "vehicle").

To estimate the distances of objects from the ego-vehicle, we employ the Video-Depth-Anything model (*36*), a state-of-the-art monocular depth estimator designed for video input. For each frame III, the model generates a dense depth map D (**Figure 13 (b)(d)**), where each pixel D(x,y) represents the predicted distance (in meters or relative units) from the camera to the object surface.

**Equation 2:** Within the segmented region R, the object's estimated distance is computed as the average depth value:

$$Z = \left(\frac{1}{|R|} \times \sum D(x, y)\right), where\ (x, y) \in R \tag{2}$$

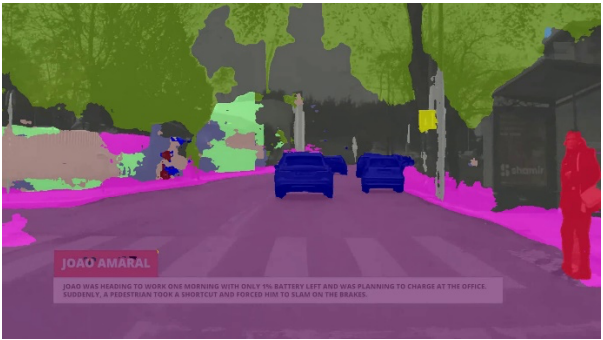

**(a) Pedestrian Segmentation Example**

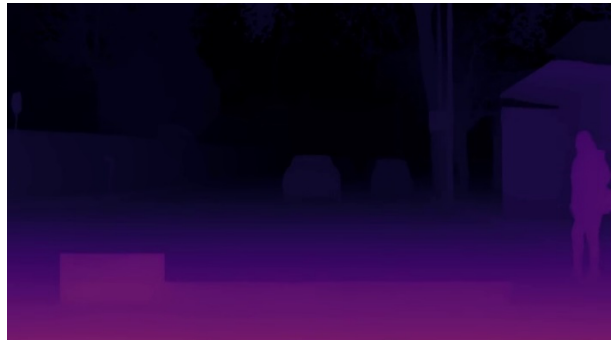

**(b) Depth Map from Monocular Estimation**

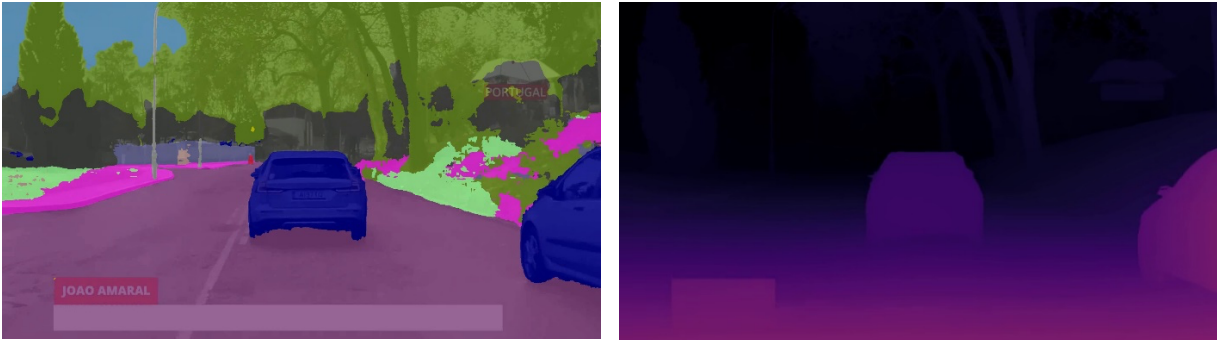

**(c) Vehicle Segmentation Example**     **(d) Depth Map from Monocular Estimation**

**Figure 13. Depth & Semantic Masks**

This study estimates Time-To-Collision (TTC) by calculating the relative velocity of a detected object based on depth changes between consecutive frames. **Equation 3:** Let f be the video frame rate (frames per second) and let Z and Z' denote the object's estimated distances at the current and next frames, respectively. The time interval is:

$$\Delta t = \frac{1}{f} \tag{3}$$

**Equation 4:** The relative velocity is:

$$V = (Z - Z')/\Delta t \tag{4}$$

TTC is defined as:

$$TTC = \frac{Z}{\Delta V} = \frac{Z \times \Delta V}{Z - Z'} \tag{5}$$

**Equation 5:** When $\Delta V > 0$, the object is approaching the camera and TTC is valid. If $\Delta V \leq 0$, the object is stationary or moving away, and TTC is set to infinity. To mitigate the effects of noise in depth estimation, we optionally apply median filtering within the object mask or use temporal smoothing across frames. In multi-object scenes, we apply connected components analysis to isolate each object instance and compute its average depth individually. The TTC is then estimated for the nearest approaching object only.

**Figure 14** illustrates an example of an evasive maneuver from the SAVeD. The blue curve shows the evolution of Time-to-Collision (TTC) over video frames. Initially, the TTC is large as the scene is calm (left image). A sudden red-light violation by an oncoming vehicle causes TTC to sharply drop below 1.5 seconds (middle image). Subsequently, the ADAS-equipped vehicle performs a right-turn evasive maneuver, increasing TTC and resolving the risk (right image).

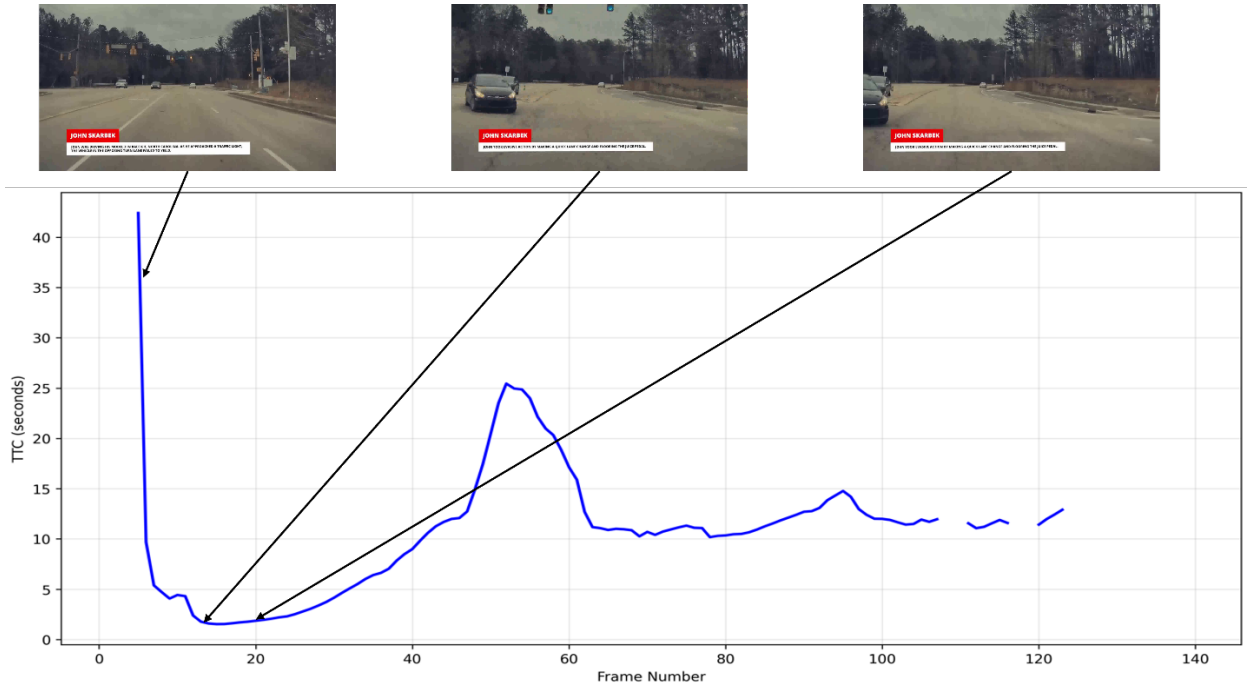

**Figure 14. Time-to-Collision Profile for Near-Miss Event**

**Table 4** summarizes the estimated parameters of the Generalized Extreme Value (GEV) distributions fitted to Time-to-Collision (TTC) observations for both crash and near-miss events extracted from the SAVeD dataset. The analysis covers three roadway types including highways, general segments, and signalized intersections, to characterize extreme risk conditions across different traffic environments. The GEV parameters $(\xi, \sigma, \mu, p)$ quantify the tail behavior of the TTC distribution, where a smaller location parameter $(\mu)$ and more negative shape parameter $(\xi)$ indicate higher concentrations of extremely low TTCs and, consequently, greater collision likelihood.

**TABLE 4 Estimated Parameters of Generalized Extreme Value (GEV) Distributions for Crash and Near-Miss Events across Roadway Types**

| Roadway Type | crash | | | | Near miss | | | |
|---|---|---|---|---|---|---|---|---|
| | $\xi$ | $\sigma$ | $\mu$ | $p$ | $\xi$ | $\sigma$ | $\mu$ | $p$ |
| Highway | -0.5580 | 1.0677 | -1.8 | 0.63 | -0.7776 | 1.7391 | -2.5 | 0 |
| Segment | -0.6385 | 1.5299 | -2.3 | 0.65 | -0.8330 | 2.1342 | -3.0 | 0 |
| Signalized | -0.5628 | 0.9253 | -1.6 | 0.16 | -0.7429 | 2.1244 | -3.1 | 0 |
| Three class | -0.6165 | 1.1994 | -1.9 | 0.23 | -0.7136 | 1.6772 | -2.6 | 0 |

**Figure 15** illustrate an application of the SAVeD dataset for modeling extreme risk conditions using the Generalized Extreme Value (GEV) distribution(*37*). The fitted GEV curves demonstrate clear behavioral distinctions between crash and near-miss cases. Crash distributions exhibit heavier left tails and lower mean TTCs, implying that ADAS-equipped vehicles in these situations experience abrupt and severe reductions in temporal safety margins. Conversely, the near-miss distributions shift rightward with larger $\mu$ and $\sigma$ parameters, corresponding to longer TTCs and more successful evasive maneuvers.

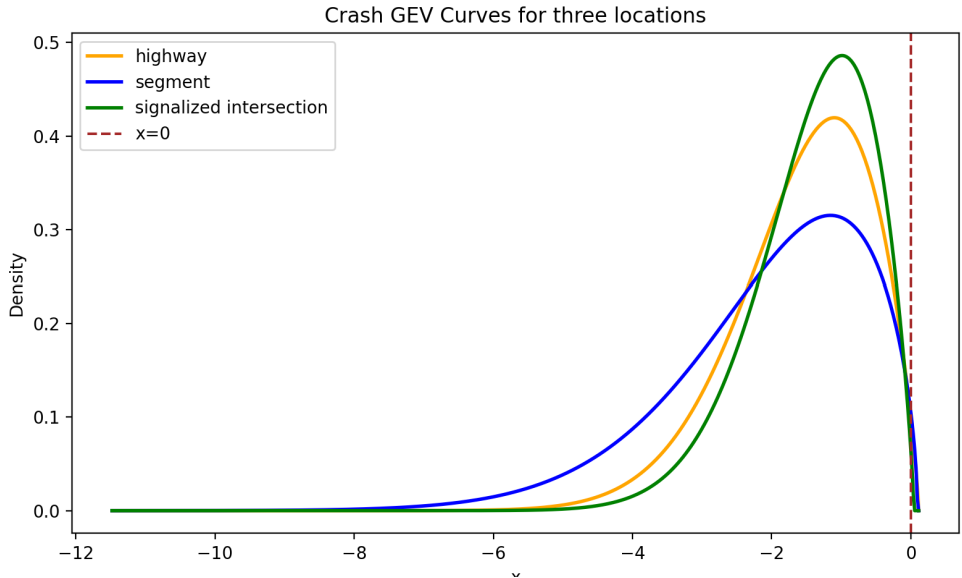

**(a) The GEV curves for crashes by three locations based on EVT**

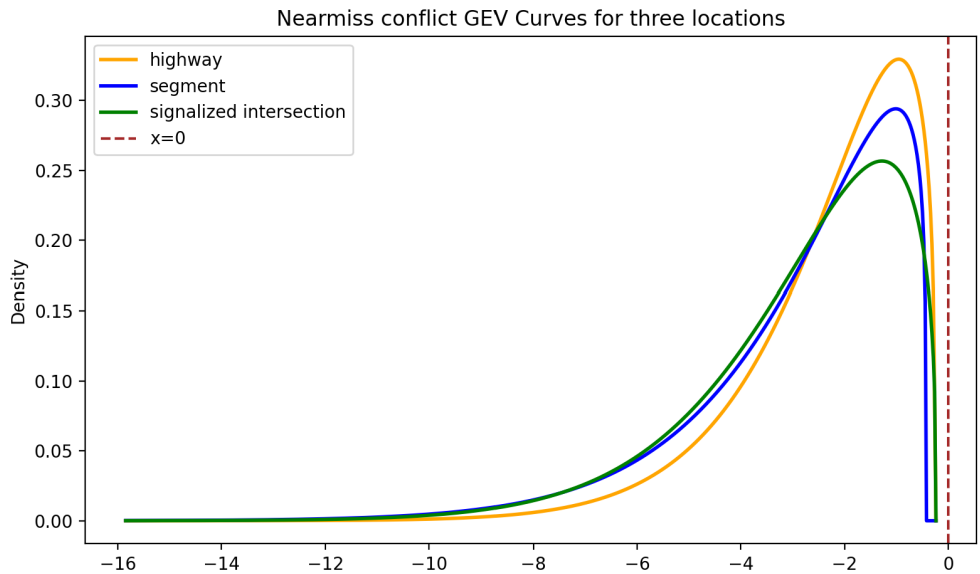

**(a) The GEV curves for Near misses conflicts by three locations based on EVT**
**Figure 15. GEV curves for crashes and Near misses based on EVT**

Among the three roadway contexts, signalized intersections show the steepest TTC decay and highest density near extreme values, reflecting the complexity of stop-and-go conditions and cross-traffic interactions. Highways show smoother TTC profiles, while general road segments lie between the two extremes.

These findings demonstrate that extreme-value modeling provides a robust framework for quantifying near-collision risks and differentiating successful avoidance events from actual crashes. By integrating computer-vision-derived TTC measures with GEV analysis, the SAVeD dataset enables systematic evaluation of ADAS evasive performance under diverse roadway conditions and supports probabilistic risk assessment for future intelligent-vehicle systems.

**Technical validation**

Currently, most first-person accident datasets lack detailed and structured annotations. Existing video-based datasets such as DADA-2000 (*6*), CCD (36), and AV-TAU (*38*) primarily provide first-person or CCTV-style traffic accident footage(*39–43*), but their annotations are often limited to single-sentence event descriptions or question–answer pairs designed for training large language models (LLMs) rather than for systematic traffic safety analysis(*44–46*).

For example, these datasets typically describe an event with brief captions such as "a pedestrian crosses and a collision occurs," without specifying key contextual attributes like vehicle behavior, environmental factors, risk evolution, or driver attention. Consequently, they offer only coarse-grained supervision signals and lack the depth needed for quantitative safety assessment.

In contrast, the SAVeD dataset introduces fine-grained, semantically rich annotations covering temporal, spatial, and behavioral dimensions, making it a valuable resource for traffic safety research and proactive risk modeling. Moreover, the detailed contextual information in SAVeD can also benefit LLM training, as it provides richer scene understanding cues and temporal reasoning structure, bridging the gap between perception-level video data and high-level safety interpretation.

Utilizing the detailed richness of the SAVeD, we proceed to establish and evaluate baseline models to serve as a strong reference for subsequent studies. We report baselines on SAVeD with VideoLLaMA2 and InternVL2.5 HiCo R16. Exact match is used for categorical fields and cosine similarity for text fields. We fine tuned InternVL2.5 HiCo R16 on eighty percent of the data and evaluated on the remaining twenty percent.

**Table 5 Macro and weighted averages by split**

| Split | Categorical macro | Text macro | Overall macro | Categorical weighted | Text weighted | Overall weighted |
|-------|-------------------|------------|---------------|----------------------|---------------|------------------|
| Near miss baseline InternVL2.5 HiCo R16 (*47*) | 26.02 | 29.00 | 27.02 | 26.02 | 29.00 | 27.02 |
| Near miss baseline VideoLLaMA 2 (*48*) | 9.39 | 26.66 | 15.14 | 9.39 | 26.66 | 15.14 |
| Crash baseline VideoLLaMA 2 | 58.90 | 77.42 | 67.77 | 62.50 | 77.42 | 67.77 |
| Crash test fine tuned InternVL2.5 HiCo R16 | 60.90 | 39.90 | 53.49 | 65.01 | 40.09 | 55.97 |
| Near miss test fine tuned InternVL2.5 HiCo R16 | 76.25 | 44.77 | 65.76 | 76.25 | 44.95 | 65.90 |

As shown in **Table 5**, the near-miss baselines show clear regularities. VideoLLaMA2 reaches 0.68 in light condition, 33.96 in weather, 0.51 in road surface condition, zero in road type and road flat, and 21.16 in lanes. Its text alignment is 35.13 in traffic condition, 16.10 in guilty, and 28.75 in avoid reason. InternVL2.5 HiCo R16 improves weather to 47.27, road surface condition to 84.81, and road flat to 22.70, while remaining at 0.34 for road type and 1.02 for lanes. Text alignment rises modestly to 36.97 in traffic condition, 21.26 in guilty, and 28.79 in avoid reason. Crash baselines with VideoLLaMA2 present a stronger profile. Light condition is 87.65, weather is 77.84, road surface condition is 91.27, and road flat is 98.53. Road type is 39.51 and lanes is 15.88. Crash-level attributes are 26.86 for crash type, 36.86 for type of impact, 43.24 for crash vehicle type, 77.06 for total number of vehicles, and 92.84 for crash severity. In text, traffic condition is 55.32, ego pre-avoidance movement is 93.35, other pre-avoidance movement is 97.34, guilty is 95.66, crash reason is 44.36, and other most damaged area is 78.51. We then fine-tuned InternVL2.5 HiCo R16 and measured performance on the held-out test set. On crash scenes the model achieves 87.75 in light condition, 78.43 in weather, 92.16 in road surface condition, and 99.02 in road flat. Road type is 50.49 and lanes is 27.94. Crash-level attributes are 34.31 for crash type, 35.78 for type of impact, zero for crash vehicle type, 73.04 for total number of vehicles, and 90.95 for crash severity. Text alignment is 66.23 in traffic condition, 25.28 in ego pre-avoidance movement, 24.15 in other pre-avoidance movement, 41.96 in guilty, 23.45 in crash reason, and 58.31 in other most damaged area. On near-miss scenes after fine-tuning the model attains 85.84 in light condition, 69.03 in weather, 91.15 in road surface condition, 65.49 in road type, 97.35 in road flat, and 48.67 in lanes. In text it reaches 74.58 in traffic condition, 23.39 in guilty, and 36.33 in avoid reason. These outcomes show that domain adaptation improves both recognition and description in the visually variable near-miss setting.

### Summary

This study provides a comprehensive analysis of ADAS-equipped vehicle-involved crashes and near-miss events using the SAVeD dataset, which offers rich macro-level annotations and visual evidence. The contributions of current study are summarized as follows:

**1. Empirical insights into ADAS performance under real-world conditions.**

By systematically extracting variables in 2,119 first-person videos such as lighting, weather, road type, crash type, damaged areas, repair cost estimates, and driver assistance levels, we reveal critical patterns in ADAS-equipped vehicle performance and risk exposure under varying conditions. The findings show that ADAS-equipped vehicles are more prone to fault attribution in complex weather and road environmental scenarios—such as poor lighting, wet surfaces, and intersections—while demonstrating stronger evasive capabilities when supervised autonomy is engaged.

**2. Introduction of a novel, publicly available video dataset.**

SAVeD stands as the one of the first large-scale, real-world dataset documenting ADAS-equipped vehicle crashes and near-miss events. Its fine-grained temporal annotations, documentation of safety driver behavior, and multi-modal richness—including user-generated public sentiment—enable diverse applications: from evaluating rare edge-case scenarios, to benchmarking planning and prediction models, to supporting human-machine collaboration research and vision-language modeling.

**3. Foundation for interpretable and robust ADAS development.**

While this study focuses on dataset construction and preliminary analysis, SAVeD fills a critical data gap and lays the foundation for future research in simulation, crash prediction, driver intent modeling and real-world safety evaluation.

Future research may build on this dataset to explore simulation, prediction, and decision-making models under real-world complexities.

### Data availability

To reflect advances in ADAS technology, SAVeD is updated twice a year with new videos covering diverse countries, weather, and time conditions. Each release includes detailed notes to support longitudinal performance analysis. Community contributions are encouraged via GitHub and social media. Submissions undergo privacy and quality review and may be included in future releases to enhance the dataset's scale and diversity.

For convenience, researchers may use open-source tools such as youtube-dl(*49*) to download publicly available videos, while ensuring compliance with all applicable platform guidelines. To extract only the annotated portions of each video, we also recommend the use of YouTubeVideoSegmentDownloader(*50*), a lightweight Python tool that allows users to download specific time intervals from YouTube videos.

This open-access release is designed to support reproducibility, community benchmarking, and responsible research in ADAS-equipped vehicle safety analysis.

### Code Availability Statement

All custom annotation and data-processing code used in this study is openly available at the SAVeD GitHub repository (https://github.com/ShaoyanZhai2001/SAVeD). Code used for the Technical Validation analyses is provided in a dedicated subdirectory at: https://github.com/ShaoyanZhai2001/SAVeD/tree/main/crash_nearmiss_analyzer. No proprietary code was used.

### Author Contributions

S.Z., M.A.-A., and C.W. designed the study and the dataset.
S.Z., M.A.-A., and C.W. collected and annotated the data.
S.Z., M.A.-A., and C.W. implemented the analysis scripts.
S.Z., M.A.-A., C.W., and R.V.G. performed the technical validation.
S.Z., M.A.-A., C.W., and R.V.G. drafted the manuscript.
All authors reviewed and approved the final manuscript.

### Competing Interests

The authors declare no competing interests.

### Funding: Not applicable.

### Ethics Statement

All videos in the SAVeD dataset are sourced from publicly accessible social media platforms and are used solely for academic research and education. Sensitive content involving injury to people or animals has been removed to comply with ethical standards and platform policies. No identifiable personal information is included in the dataset. Original video files are not redistributed; only URLs and annotated start and end timestamps are provided, and users must download the videos themselves in accordance with platform terms of service.